\documentclass[runningheads,a4paper]{llncs}

\usepackage{natbib}

\usepackage{xcolor}
\usepackage[font={footnotesize,sf}]{caption}

\makeatletter
\renewenvironment{quotation}
               {\list{}{\listparindent=0pt%whatever you need
                        \itemindent    \listparindent
                        \leftmargin=0pt%  whatever you need
                        \rightmargin=10pt%whatever you need
                        \topsep=0pt%%%%%  whatever you need
                        \parsep        \z@ \@plus\p@}%
                \item\relax}
               {\endlist}
\makeatother

\newcommand{\mybox}[2][l]{%
  \colorbox{white}{\renewcommand{\arraystretch}{0}\begin{tabular}{@{}#1@{}}\color{colorFilm!80!black}\tiny\bf\sffamily#2\end{tabular}}%
}

\newcommand{\listingref}[1]{Listing {\small #1}}
\newcommand{\listinglabel}[1]{\label{listing:#1}}

\newcommand{\imageschema}[1]{{\sffamily\textsc{\lowercase{#1}}}}

\usepackage{times}
\usepackage{url}
\usepackage{latexsym}

\usepackage[pdftex]{graphicx}
\usepackage{tabularx}

\definecolor{mathcolor}{RGB}{7,72,110}
\definecolor{baseColor}{RGB}{0,97,161}
\definecolor{colorFilm}{RGB}{0,97,161}
\usepackage{color,xspace}
\usepackage{setspace}

\definecolor{DarkBlue}{RGB}{7,72,110}
\definecolor{LightBlue}{RGB}{205,237,249}
\definecolor{LightRed}{RGB}{255,102,102}
\definecolor{DarkRed}{RGB}{153,0,0}
\definecolor{LightGreen}{RGB}{178,205,102}
\definecolor{DarkGreen}{RGB}{0,51,0}

\usepackage{amssymb}

\usepackage[geometry]{ifsym}
\newcommand{\mylabel}[1]{{\footnotesize\bf({\color{blue!50!black}\footnotesize#1}).}}
\newcommand{\bul}{$\color{blue!80!black}\blacktriangleright~$}
\newcommand{\marker}{\color{blue}\tiny\FilledSquareShadowA}

\newcommand{\thePipe}[0]{$~$\rule[-0.4ex]{0.3ex}{1em}$~$ }

\newcommand{\analysisSkip}[0]{\smallskip}
\newcommand{\analysisOneUnit}[2]{\tiny #1}
\newcommand{\spatial}[1]{{\textsc{#1}}}
\newcommand{\imageryObject}[1]{\sffamily{\color{colorFilm}#1}}
\newcommand{\imageryIdentity}[1]{\emph{#1}}

\newcommand{\narrativeTextFilm}[7]{
\begin{center}
\colorbox{colorFilm!80!black}{
	\begin{minipage}{0.98\columnwidth}
		{\color{white}\scriptsize\sffamily \uppercase{{\bf #1}} \thePipe \uppercase{{#3}}.\\[-4pt]{Director}.\quad #2}
	\end{minipage}}
\colorbox{gray!18}{
	\begin{minipage}{0.98\columnwidth}
		{\scriptsize\sffamily #4}
	\end{minipage}}
\colorbox{colorFilm!80!black}{
	\begin{minipage}{0.98\columnwidth}
		{\color{white}\tiny\sffamily \textbf{Credit}.\quad #5}\hfill \mybox[c]{#7\listinglabel{#7}}
	\end{minipage}}
\end{center}
}

\newcommand{\filmCharacter}[1]{\sffamily{\color{colorFilm}#1}}
\newcommand{\filmCinematography}[1]{{\color{colorFilm}\textsc{#1}}}
\newcommand{\filmIdentity}[1]{\emph{#1}}

\newcommand{\narrativeTextDesign}[7]{
\begin{center}
\colorbox{colorFilm!80!black}{
	\begin{minipage}{0.98\columnwidth}
		{\color{white}\scriptsize\sffamily \uppercase{{\bf #1}} \thePipe \uppercase{{#3}}.\\[-4pt]{Location}.\quad #2}
	\end{minipage}}
\colorbox{gray!18}{
	\begin{minipage}{0.98\columnwidth}
		{\scriptsize\sffamily #4}
	\end{minipage}}
\colorbox{colorFilm!80!black}{
	\begin{minipage}{0.98\columnwidth}
		{\color{white}\tiny\sffamily \textbf{Credit}.\quad #5}\hfill \mybox[c]{#7\listinglabel{#7}}
	\end{minipage}}
	
\end{center}
}

\usepackage{array}
\newcolumntype{L}[1]{>{\raggedright\let\newline\\\arraybackslash\hspace{0pt}}m{#1}}
\newcolumntype{C}[1]{>{\centering\let\newline\\\arraybackslash\hspace{0pt}}m{#1}}
\newcolumntype{R}[1]{>{\raggedleft\let\newline\\\arraybackslash\hspace{0pt}}m{#1}}

\newcommand{\ImageSchema}[1]{{\footnotesize\sffamily\uppercase{#1 }}}

\newcommand{\Above}[0]{\ImageSchema{Above}}
\newcommand{\Across}[0]{\ImageSchema{Across}}
\newcommand{\Covering}[0]{\ImageSchema{Covering}}
\newcommand{\Contact}[0]{\ImageSchema{Contact}}
\newcommand{\VerticalOrientation}[0]{\ImageSchema{Vertical\_Orientation}}
\newcommand{\Length}[0]{\ImageSchema{Length}}

\newcommand{\Containment}[0]{\ImageSchema{Containment}}
\newcommand{\Path}[0]{\ImageSchema{Path}}
\newcommand{\PathGoal}[0]{\ImageSchema{Path\_Goal}}
\newcommand{\SourcePathGoal}[0]{\ImageSchema{Source\_Path\_Goal}}
\newcommand{\Blockage}[0]{\ImageSchema{Blockage}}
\newcommand{\CenterPeriphery}[0]{\ImageSchema{Center\_Periphery}}
\newcommand{\Cycle}[0]{\ImageSchema{Cycle}}
\newcommand{\CyclicClimax}[0]{\ImageSchema{Cyclic\_Climax}}

\newcommand{\Compulsion}[0]{\ImageSchema{Compulsion}}
\newcommand{\Counterforce}[0]{\ImageSchema{Counterforce}}
\newcommand{\Diversion}[0]{\ImageSchema{Diversion}}
\newcommand{\RemovalOfRestraint}[0]{\ImageSchema{Removal\_of\_Restraint}}
\newcommand{\Enablement}[0]{\ImageSchema{Enablement}}
\newcommand{\Attraction}[0]{\ImageSchema{Attraction}}
\newcommand{\Link}[0]{\ImageSchema{Link}}
\newcommand{\Scale}[0]{\ImageSchema{Scale}}

\newcommand{\AxisBalance}[0]{\ImageSchema{Axis\_Balance}}
\newcommand{\PointBalance}[0]{\ImageSchema{Point\_Balance}}
\newcommand{\TwinPanBalance}[0]{\ImageSchema{Twin\_Pan\_Balance}}
\newcommand{\Equilibrium}[0]{\ImageSchema{Equilibrium}}

\newcommand{\LinearPathFromMovingObject}[0]{\ImageSchema{Linear\_Path\_from\_Moving\_Object}}
\newcommand{\PathToEndpoint}[0]{\ImageSchema{Path\_to\_Endpoint}}
\newcommand{\PathToObjectMass}[0]{\ImageSchema{Path\_to\_Object\_Mass}}
\newcommand{\MultiplexToMass}[0]{\ImageSchema{Multiplex\_to\_Mass}}
\newcommand{\Reflexive}[0]{\ImageSchema{Reflexive}}
\newcommand{\Rotation}[0]{\ImageSchema{Rotation}}

\newcommand{\Surface}[0]{\ImageSchema{Surface}}
\newcommand{\FullEmpty}[0]{\ImageSchema{Full--Empty}}
\newcommand{\Merging}[0]{\ImageSchema{Merging}}
\newcommand{\Matching}[0]{\ImageSchema{Matching}}
\newcommand{\NearFar}[0]{\ImageSchema{Near--Far}}
\newcommand{\MassCount}[0]{\ImageSchema{Mass--Count}}
\newcommand{\Iteration}[0]{\ImageSchema{Iteration}}
\newcommand{\Object}[0]{\ImageSchema{Object}}
\newcommand{\Splitting}[0]{\ImageSchema{Splitting}}
\newcommand{\PartWhole}[0]{\ImageSchema{Part-Whole}}
\newcommand{\Superimposition}[0]{\ImageSchema{Superimposition}}
\newcommand{\Process}[0]{\ImageSchema{Process}}
\newcommand{\Collection}[0]{\ImageSchema{Collection}}

\newcommand{\location}[1]{{\sffamily{\color{blue!60!gray}\textsc{#1}}}}
\newcommand{\action}[1]{{\sffamily{\color{red!70!blue}\textsc{#1}}}}

\newcommand{\timepoint}[0]{{\color{blue!60!gray}\bullet}}

\newcommand{\sentence}[1]{{\begin{center}\colorbox{gray!7}{\begin{minipage}{0.96\columnwidth} \centering\sffamily\footnotesize{#1}\end{minipage}}\end{center}}}

\begin{document}

\mainmatter  % start of an individual contribution

\title{{\Huge\sffamily \textsc{Talking about}\\[7pt]\textsc{the Moving Image}} \\$~$\\ {\large\sffamily A Declarative Model for Image Schema Based\\ Embodied Perception Grounding and Language Generation}}

%\title{{\huge\sffamily Deep Semantics for\\[5pt]the Moving Image} \\$~$\\ {\large\sffamily A Declarative Model for Image Schema Based\\ Embodied Perception Grounding and Language Generation}}

\titlerunning{\textsc{Talking about the Moving Image}}

\author{ Jakob Suchan\inst{1,2} \and Mehul Bhatt\inst{1,2} \and Harshita Jhavar\inst{2,3}
}
\authorrunning{}


\institute{University of Bremen, Germany\\[4pt]
\and
The DesignSpace Group\\\url{www.design-space.org/Next}\\[4pt]
\and MANIT (Bhopal, India)
}

\toctitle{Spatial Symmetry Based Pruning}
\tocauthor{}
\maketitle

\begin{abstract}
We present a general theory and corresponding declarative model for the embodied grounding and natural language based analytical summarisation of dynamic visuo-spatial imagery. The declarative model ---ecompassing spatio-linguistic abstractions, image schemas, and a spatio-temporal feature based language generator--- is modularly implemented within Constraint Logic Programming (CLP). The implemented model is such that primitives of the theory, e.g., pertaining to space and motion, image schemata, are available as first-class objects with \emph{deep semantics} suited for inference and query. We demonstrate the model with select examples broadly motivated by areas such as \emph{film, design, geography}, \emph{smart environments} where analytical natural language based externalisations of \emph{the moving image} are central from the viewpoint of human interaction, evidence-based qualitative analysis, and sensemaking.

\keywords{\emph{moving image, visual semantics and embodiment, visuo-spatial cognition and computation, cognitive vision, computational models of narrative, declarative spatial reasoning}}

\end{abstract}

% \tableofcontents 

\medskip

\section{\textsc{Introduction}}\label{sec:introduction}

Spatial thinking, conceptualisation, and the verbal and visual (e.g., gestural, iconic, diagrammatic) communication of commonsense as well as expert knowledge about the world ---the \emph{space} that we exist in--- is one of the most important aspects of everyday human life \citep{Tversky2005-visuospatial-reasoning-cambridge,Tversky-2004-narrative,Bhatt-sensemaking-narrative-2013}. Philosophers, cognitive scientists, linguists, psycholinguists, ontologists, information theorists, computer scientists, mathematicians have each investigated \emph{space} through the perspective of the lenses afforded by their respective field of study \citep{SC-ECAI-2004,spatial-foundations-cognition-lang-2009,bateman-space-language-ontology-2010,Bhatt:RSAC:2012,Bhatt-Schultz-Freksa:2013,hdbk-soat-cog-2013}. Interdisciplinary studies on visuo-spatial cognition, e.g., concerning `visual perception', `language and space', `spatial memory', `spatial conceptualisation', `spatial representations',  `spatial reasoning'  are extensive. In recent years, the fields of \emph{spatial cognition and computation}, and \emph{spatial information theory} have established their foundational significance for the design and implementation of computational cognitive systems, and multimodal interaction \& assistive technologies, e.g., especially in those areas where \emph{processing and interpretation} of potentially large volumes of highly \emph{dynamic spatio-temporal data} is involved \citep{Bhatt-sensemaking-narrative-2013}: cognitive vision \& robotics, geospatial dynamics \citep{GeospatialDynamics-2014}, architecture design \citep{KR-2014-Bhatt} to name a few prime examples. 

Our research addresses `\emph{space and spatio-temporal dynamics}' from the viewpoints of visuo-spatial cognition and computation, computational cognitive linguistics, and formal representation and computational reasoning about space, action, and change. We especially focus on space and motion as interpreted within artificial intelligence and knowledge representation and reasoning (KR) in general, and \emph{declarative spatial reasoning} \citep{clpqs-2011,clpqs-ecai-2012,aspmtqs-lpnmr-2015} in particular. Furthermore, the concept of \emph{image schemas} as ``\emph{abstract recurring patterns of thought and perceptual experience}'' \citep{johnson1990body,lakoff1990women} serves a central role in our formal framework.

%\begin{itemize}
%	\item visuo-spatial cognition and computation, 
%	\item computational cognitive linguistics, and 
%	\item formal representation and computational reasoning about space, action, and change 
%\end{itemize}

\subsubsection*{Visuo-Spatial Dynamics of the Moving Image}
%\subsubsection*{The Moving Image, and Visuo-Spatial Dynamics}

\emph{The Moving Image}, from the viewpoint of this paper, is interpreted in a broad sense to encompass:

{\small
\begin{quotation}
\sffamily{\textbf{multi-modal} visuo-auditory perceptual signals (also including depth sensing, haptics, and empirical observational data) where basic concepts of semantic or content level coherence, and spatio-temporal continuity and narrativity are applicable.\hfill\marker}
\end{quotation}
}

As examples, consider the following:

{%\small
\bul\emph{cognitive studies of film} aimed at investigating attention and recipient effects in observers vis-a-vis the motion picture \citep{CogMediaTheory:2014,Storytelling-cogsci-neuro-Aldama:2015}

\smallskip

\bul\emph{evidence-based design} \citep{hamilton2009evidence,cama2009evidence} involving analysis of post-occupancy user behaviour in buildings, e.g., pertaining visual perception of signage 

\smallskip

\bul\emph{geospatial dynamics} aimed at human-centered interpretation of (potentially large-scale) geospatial satellite and remote sensing imagery \citep{GeospatialDynamics-2014}

\smallskip

\bul\emph{cognitive vision and control} in robotics, smart environments etc, e.g., involving human activity interpretation and real-time object / interaction tracking in professional and everyday living (e.g., meetings, surveillance and security at an airport) \citep{Vernon2006,Vernon2008,ILP-2011-Dubba,CMN-2013-Bhatt,suchan-embodied-pricai2014,JAIR-ILP-Leeds-Bremen-2015}.
}
%
%\begin{itemize}
%	\item  \emph{cognitive studies of film} aimed at investigating attention and recipient effects in observers vis-a-vis the moving image
%	\item \emph{evidence-based architecture design} involving analysis of post-occupancy user behaviour in buildings pertaining to, for instance, eye-tracking and egocentric video capture based analysis of visual perception and attention, indoor people-movement analysis
%	\item \emph{geospatial dynamics} aimed at human-centered interpretation of (potentially large-scale) geospatial satellite imagery
%	\item \emph{cognitive vision and control} in robotics, smart environments etc, e.g., involving human activity interpretation and real-time object / interaction tracking
%\end{itemize}

Within all these areas, high-level semantic interpretation and qualitative analysis of the moving image requires the representational and inferential mediation of (declarative) embodied, qualitative abstractions of the visuo-spatial dynamics, encompassing \emph{space, time, motion, and interaction}.

%--- that are identifiable the moving image.

% (\textbf{a}--\textbf{d})
%(i.e., visuo-spatial, auditory etc sensory input) 

\subsubsection*{Declarative Model of Perceptual Narratives}
With respect to a broad-based understanding of the moving image (as aforediscussed), we define visuo-spatial \emph{perceptual narratives} as:

{\small
\begin{quotation}
\sffamily{\textbf{declarative models} of visual, auditory, haptic and other (e.g., qualitative, analytical) observations in the real world that are obtained via artificial sensors and / or human input.\hfill\marker}
\end{quotation}
}

Declarativeness denotes the existence of grounded (e.g., symbolic, sub-symbolic) models coupled with \textbf{deep semantics} (e.g., for spatial and temporal knowledge) and systematic formalisation that can be used to perform reasoning and query answering, embodied simulation, and relational learning.\footnote{Broadly, we refer to methods for abstraction, analogy-hypothesis-theory formation, belief  revision, argumentation.} With respect to methods, this paper particularly alludes to declarative KR frameworks such as logic programming, constraint logic programming, description logic based spatio-terminological reasoning, answer-set programming based non-monotonic (spatial) reasoning, or even other specialised commonsense reasoners based on expressive action description languages for handling \emph{space, action, and change}. Declarative representations serve as basis to externalise explicit and \emph{inferred} knowledge, e.g., by way of modalities such as visual and diagrammatic representations, natural language, etc.

\textbf{Core Contributions}.\quad We present a declarative model for the embodied grounding of the visuo-spatial dynamics of the moving image, and the ability to generate corresponding textual summaries that serve an analytical function from a computer-human interaction viewpoint in a range of cognitive assistive technologies and interaction system where reasoning about space, actions, change, and interaction is crucial. The overall framework encompasses: 

\mylabel{F1} a formal theory of qualitative characterisations of \emph{space and motion} with deep semantics for spatial, temporal, and motion predicates

\smallskip

\mylabel{F2} formalisation of the embodied \emph{image schematic} structure of visuo-spatial dynamics wrt. the formal theory of space and motion

\smallskip

\mylabel{F3} a declarative \emph{spatio-temporal feature-based natural language generation engine} that can be used in a domain-independent manner

%implementing the full pipeline (from text planing to realisation)

The overall framework ({\small\color{blue!50!black}\bf F1--F3}) for the embodied grounding of the visuo-spatial dynamics of the moving image, and the externalisation of the declarative perceptual narrative model by way of natural language has been fully modelled and implemented in an elaboration tolerant manner within Constraint Logic Programming (CLP). We emphasize that the level of declarativeness within logic programming is such that each aspect pertaining to the overall framework can be seamlessly customised and elaborated, and that question-answering \& query can be performed with spatio-temporal relations, image schemas, path \& motion predicates, syntax trees etc as first class objects within the CLP environment.

\textbf{Organization of the Paper}.\quad Section \ref{sec:talking-about-sm} presents the application scenarios that we will directly demonstrate as case-studies in this paper; we focus on a class of cognitive interaction systems where the study of visuo-spatial dynamics in the context of the moving image is central. Sections \ref{sec:space_time_motion}--\ref{sec:image-schemas} present the theory of space, motion, and image schemas elaborating on its formalisation and declarative implementation within constraint logic programming. Section \ref{sec:language-generation} presents a summary of the declarative natural language generation component. Section \ref{sec:discussion-and-relwork} concludes with a discussion of related work.

%Section \ref{sec:discussion-related-work} includes a discussion of related and contextualisation of related work; we summarise and present pointers to the outlook of this research in Section \ref{sec:summary-conclusion-outlook}.

%\section{\textsc{Background and Related Work}}\label{sec:background-related-work}

\section{\textsc{Talking about the Moving Image}}\label{sec:talking-about-sm}
%\section{\textsc{Talking about Space and Motion: Case Studies}}\label{sec:talking-about-sm}

%Talking about space and motion denotes 

Talking about the moving image denotes: 

{\small
\begin{quotation}
\sffamily{the ability to computationally generate semantically well-founded, embodied, multi-modal (e.g., \textbf{natural language}, iconic, diagrammatic) \textbf{externalisations} of dynamic visuo-spatial phenomena as perceived via visuo-spatial, auditory, or sensorimotor haptic interactions.\hfill\marker}
\end{quotation}
}
% in so far as grounded features from other perceptual sources are available

In the backdrop of the twin notions of \emph{the moving image} \& \emph{perceptual narratives} (Section \ref{sec:introduction}), we focus on a range of computer-human interaction systems \& assistive technologies at the interface of language, logic, and cognition; in particular, visuo-spatial cognition and computation are most central. Consider the case-studies in ({\small\bf\color{blue} S1--S4}):\footnote{The paper is confined to visual processing and analysis, and `talking about it'  by way of natural language externalisations. We emphasise that our underlying model is general, and elaboration tolerant to other kinds of input features.}

%This paper focusses on \emph{natural language externalisations} of perceptual narratives.
%\footnote{The examples in  ({\small\bf S1--S4}) presuppos }

\begin{figure}[t]
    \centering
    \includegraphics[width=0.96\textwidth]{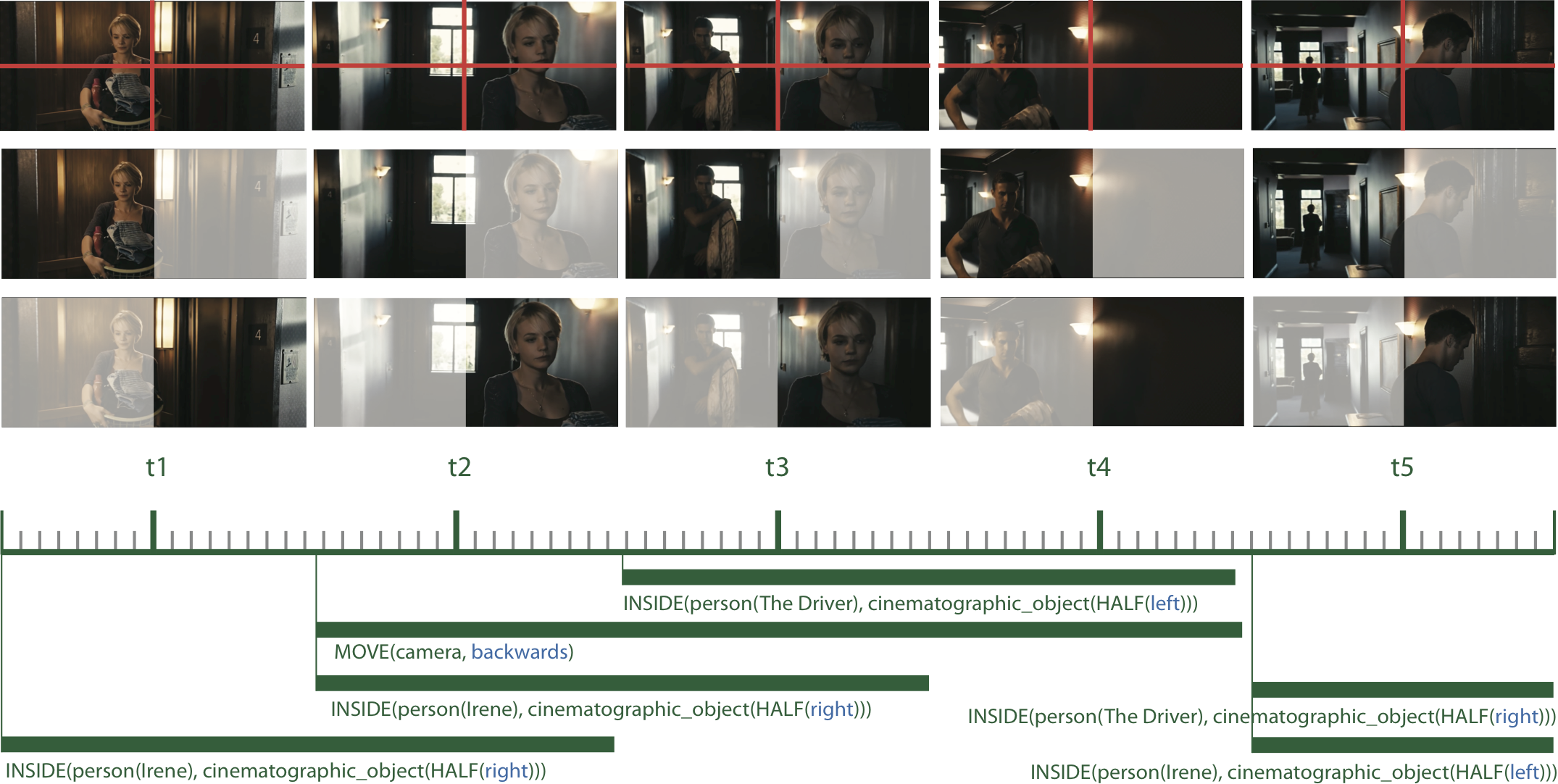}
    \caption{Analysis based on the Quadrant system (Drive 2011)}
    \label{fig:quadrant _system_method}
\end{figure}

\subsubsection*{\small \mylabel{S1}\quad \textsc{Cognitive Studies of Film}}
Cognitive studies of the moving image ---specifically, \emph{cognitive film theory}--- has accorded a special emphasis on the role of \emph{mental activity of observers} (e.g., subjects, analysts, general viewers /  spectators) as one of the most central objects of inquiry \citep{CogMediaTheory:2014,Storytelling-cogsci-neuro-Aldama:2015} (e.g., expert analysis in \listingref{L1}; Fig \ref{fig:quadrant _system_method}).  Amongst other things, cognitive film studies concern making sense of subject's visual fixation or saccadic eye-movement patterns whilst watching a film and correlating this with deep semantic analysis of the visuo-auditory data (e.g., fixation on movie characters, influence of cinematographic devices such as \emph{cuts} and sound effects on attention), studies in embodiment \citep{embodi-film-vivian-2004,Coegnarts:Image-Schema-2012}.

\begin{figure}
\narrativeTextFilm{Drive (2011)}{Nicolas Winding Refn}{Quadrant System. Visual Attention}{

	\analysisOneUnit{This short scene, involving \filmCharacter{The Driver} (\filmIdentity{Ryan Gosling}) and \filmCharacter{Irene} (\filmIdentity{Carey Mulligan}), adopts a \spatial{top-bottom} and \spatial{left-right} quadrant system that is executed in a \filmCinematography{single take} / without any \filmCinematography{cuts}}{}

	\analysisSkip
	
	\analysisOneUnit{The \filmCinematography{camera} \spatial{moves} \spatial{backward} tracking the movement of \filmCharacter{The Driver} and \filmCharacter{Irene}; \spatial{during}  \spatial{movement\_1}, \filmCharacter{Irene} \spatial{occupies} the right quadrant, \spatial{while} \filmCharacter{The Driver} \spatial{occupies} the \spatial{left} quadrant}{}

	\analysisSkip
	
	\analysisOneUnit{Spectator eye-tracking data suggests that the audience is repeatedly switching their attention between the \spatial{left} and \spatial{right} quadrants, with a majority of the audience fixating visual attention on \filmCharacter{Irene} as she \spatial{moves} into an extreme \filmCinematography{close-up shot}}{}

}{Quadrant system method based on study by Tony Zhou.}{https://www.youtube.com/watch?v=wsI8UES59TM}{L1}
\end{figure}

\subsubsection*{\small  \mylabel{S2}\quad  \textsc{Evidence Based Design (EBD) of the\\Built Environment}}
Evidence-based building design involves the study of the post-occupancy behaviour of building users with the aim to provide a scientific basis for generating best practice guidelines aimed at improving building performance and user experience. Amongst other things, this involves an analysis of the visuo-locomotive navigational experience of subjects based on eye-tracking and egocentric video capture based analysis of visual perception and attention, indoor people-movement analysis, e.g., during a wayfinding task, within a large-scale built-up environment such as a hospital or an airport (e.g., see \listingref{L2}). EBD is typically pursued as an interdisciplinary endeavour ---involving environmental psychologists, architects, technologists--- toward the development of new tools and processes for data collection, qualitative analysis etc.

%aiming is to provide a scientific basis for generating best practice guidelines aimed at improving building performance. 

\begin{figure}
\narrativeTextDesign{The New Parkland Hospital}{Dallas, Texas}{Wayfinding Study}{

	\analysisOneUnit{This experiment was conducted with 50 subjects at the New Parkland Hospital in Dallas}

	\analysisSkip
	
	\analysisOneUnit{\imageryIdentity{Subject 21} (\imageryObject{Barbara}) performed a wayfinding task (\#T5), \spatial{starting from} the reception desk of the emergency department and \spatial{finishing at}  the Anderson Pharmacy. Wayfinding task \#5 \spatial{goes through} the long corridor in the emergency department, the main reception and the blue elevators, {going up} to Level 2 \spatial{into} the Atrium Lobby, \spatial{passing through} the Anderson-Bridge, finally \spatial{arriving at} the X-pharmacy}

	\analysisSkip
	
	\analysisOneUnit{Eye-tracking data and video data analysis suggests that \imageryObject{Barbara} fixated on passerby \imageryObject{Person\_5} for two seconds as \imageryObject{Person\_5} \spatial{passes from} her \spatial{right} \spatial{in} the long corridor. \imageryObject{Barbara} fixated most \spatial{on} the big blue elevator signage \spatial{at} the main reception desk. \spatial{During} the 12th minute, video data from external GoPro cameras and egocentric video capture and eye-tracking suggest that \imageryObject{Barbara} looked indecisive (\emph{stopped walking, looked around, performed rapid eye-movements}}

}{Based on joint work with Corgan Associates (Dallas)}{http://www.design-space.org/edra45}{L2}
\end{figure}

\subsubsection*{\small  \mylabel{S3}\quad \textsc{Geospatial Dynamics}}
The ability of semantic and qualitative analytical capability to complement and synergize with statistical and quantitatively-driven methods has been recognized as important within geographic information systems. Research in geospatial dynamics \citep{GeospatialDynamics-2014} investigates the theoretical foundations necessary to develop the computational capability for high-level commonsense, qualitative analysis of dynamic geospatial phenomena within next generation event and object-based GIS systems.

\subsubsection*{\small  \mylabel{S4}\quad \textsc{Human Activity Interpretation}}
Research on embodied perception of vision  ---termed \emph{cognitive vision} \citep{Vernon2006,Vernon2008,CMN-2013-Bhatt}--- aims to enhance classical computer vision systems with cognitive abilities to obtain more robust vision systems that are able to adapt to unforeseen changes, make ``narrative'' sense of perceived data, and exhibit interpretation-guided goal directed behaviour. The long-term goal in cognitive vision is to provide general tools (integrating different aspects of space, action, and change) necessary for tasks such as real-time human activity interpretation and dynamic sensor (e.g., camera) control within the purview of vision, interaction, and robotics.

%Consider \listingref{L4} describing commonsense level people-interactions in an everyday context:
%
%
%\narrativeTextCogVision{ROTUNDE}{Smart Meeting Room. University of Bremen, DE}{Interactions. Activity Interpretation}{
%
%	\analysisOneUnit{\imageryObject{Person P} \spatial{stands up}, \spatial{passes behind} \imageryObject{person Q} \spatial{while} \spatial{moving towards} the exit and {\color{blue} leaves} the room. }{}
%
%
%}{\citepcolor{Rotunde-2013}{white}}{http://cognitive-vision.org/}{L4}
%

%To represent the continuity of spatial change, Freksa \citep{Freksa1991} introduced the \emph{conceptual neighborhoods}. Relations between two entities are conceptual neighbors if they can be directly transformed from one relation into the other by continuous change of the environment.

\section{Space, Time, and Motion}\label{sec:space_time_motion}
Qualitative Spatial \& Temporal Representation and Reasoning (QSTR) \citep{Cohn2001} abstracts from an exact numerical representation by describing the relations between objects using a finite number of symbols. Qualitative representations use a set of relations that hold between objects to describe a scene. Galton \citep{Galton1993,Galton1995,Galton2000} investigated movement on the basis of an integrated theory of space, time, objects, and position. Muller \citep{Muller1998} defined continuous change using 4-dimensional regions in space-time. Hazarika and Cohn \citep{Hazarika2002} build on this work but used an interval based approach to represent spatio-temporal primitives.  

We use spatio-temporal relations to represent and reason about different aspects of space, time, and motion in the context of visuo-spatial perception as described by \citep{Suchan2014}. 
To describe the spatial configuration of a perceived scene and the dynamic changes within it we combine spatial calculi to a general theory for declaratively reason about spatio-temporal change. 
The domain independent theory of \emph{Space}, \emph{Time}, and \emph{Motion} ($\mathsf{\Sigma}_{\mathsf{STM}}$) consists of:

\smallskip

{
\bul $\mathsf{\Sigma}_{\mathsf{Space}}$ -- {Spatial Relations} on topology, relative position, relative distance of spatial objects

\smallskip

\bul $\mathsf{\Sigma}_{\mathsf{Time}}$ -- Temporal Relations for representing relations between time points and intervals 

\smallskip

\bul$\mathsf{\Sigma}_{\mathsf{Motion}}$ -- Motion Relations on changes of distance and size of spatial objects 

\smallskip
}

The resulting theory is given as:
%
%{
%\footnotesize
%%\scriptsize
%\begin{subequations}\label{}
%\begin{align}
%\begin{split}
$\mathsf{\Sigma}_{\mathsf{STM}} \equiv_{def} [\mathsf{\Sigma}_{\mathsf{Space}} \cup \mathsf{\Sigma}_{\mathsf{Time}} \cup \mathsf{\Sigma}_{\mathsf{Motion}}]$.
%\end{split}
%\end{align}
%\end{subequations}\normalsize
%}

 \begin{figure}[t]
    \centering
    \includegraphics[width=\linewidth]{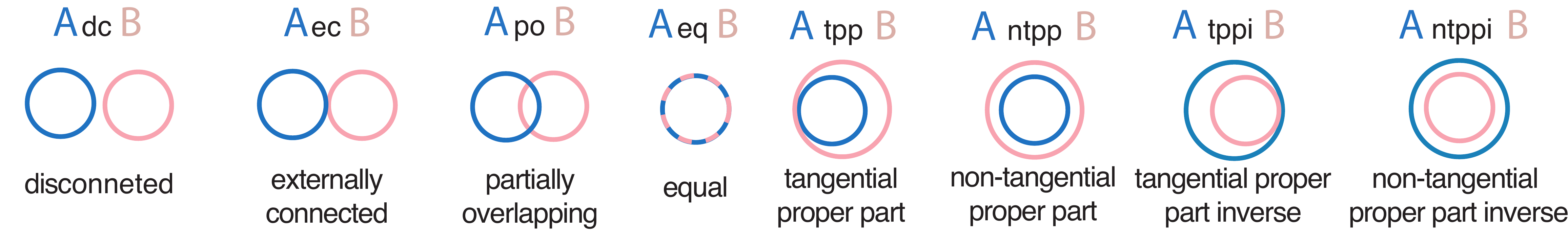}
    \caption{Region Connection Calculus (RCC-8)}
    \label{fig:rcc8}
\end{figure}

\begin{figure*}[t]
    \centering
    \includegraphics[width=0.96\textwidth]{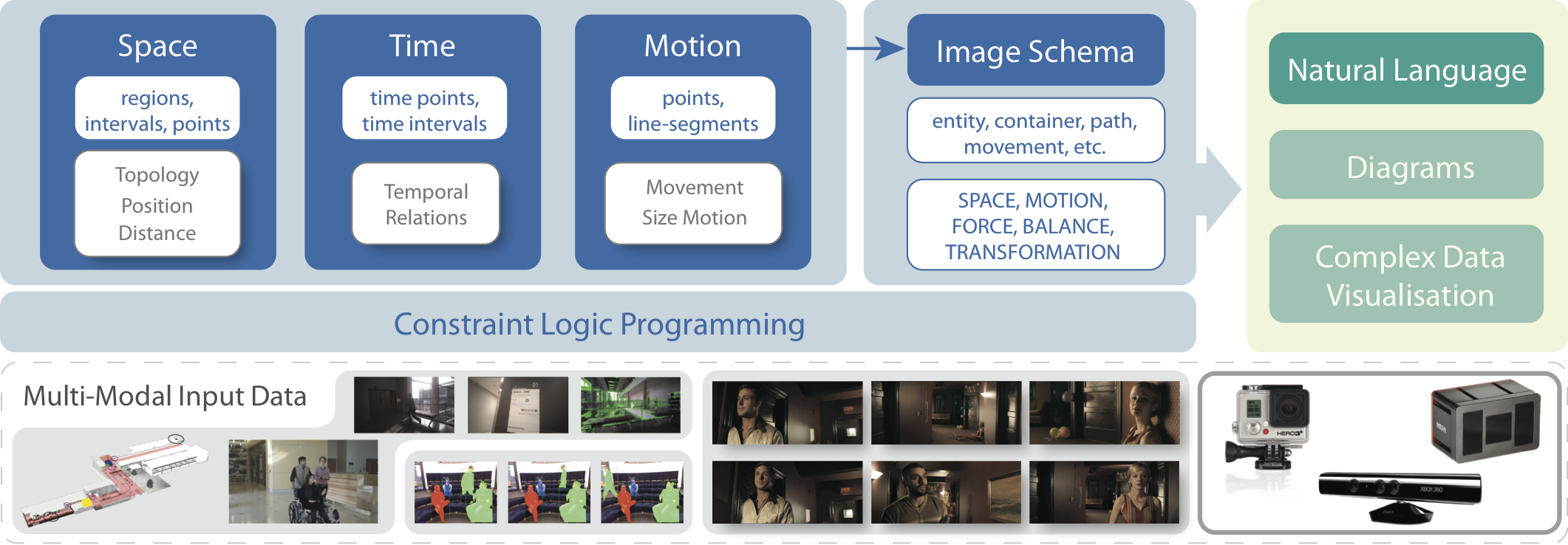}
    \caption{General Theory of Space, Time, Motion, and Image Schema}
    \label{fig:architecture-language-generation}
\end{figure*}

\smallskip
  
Objects and individuals are represented as spatial primitives according to the nature of the spatial domain we are looking at, i.e., \emph{regions of space} $\mathcal{S} = \{s_1, s_2, ..., s_n \}$,  \emph{points} $\mathcal{P} = \{p_1, p_2, ..., p_n \}$, and \emph{line segments} $\mathcal{L} = \{l_1, l_2, ..., l_n \}$ . Towards this we use functions that map from the object or individual to the corresponding spatial primitive.
%
%\medskip
%
%\emph{spatial entities} $\mathcal{E} = \{e_1, e_2, ..., e_n \}$ 
%
%\emph{extend}:  $\mathcal{E} \rightarrow \mathcal{S}$
%
%\emph{point}:  $\mathcal{E} \rightarrow \mathcal{P}$
%
%\emph{line segment}:  $\mathcal{E} \rightarrow \mathcal{P}$
%
%\medskip
%
%
%
%We use domain dependent functions that determine the spatial primitives of an object.
%
%\emph{extend}:  $\mathcal{O} \rightarrow \mathcal{S}
%
%\emph{centroid}:  $\mathcal{O} \rightarrow \mathcal{P}
%
%propositional and functional fluents
%
The spatial configuration is represented using $n$-ary \emph{spatial relations} $\mathcal{R} = \{r_1, r_2, ... ,r_n\}$ of an arbitrary spatial calculus.
$\Phi = \{\phi_1, \phi_2, ..., \phi_n\}$ is a set of propositional and functional fluents, e.g. $\phi(e_1, e_2)$ denotes the spatial relationship between $e_1$ and $e_2$. Temporal aspects are represented using \emph{time points} $\mathcal{T} = \{t_1, t_2, ..., t_n \}$  and \emph{time intervals} $\mathcal{I} = \{i_1, i_2, ..., i_n \}$.
%$Holds(\phi, r, t) \subset [\Phi, \mathcal{R}, \mathcal{T}]$
%$Holds(\phi, r, t)$
$Holds(\phi, r, at(t))$ is used to denote that the fluent $\phi$ has the value $r$ at time $t$.
To denote that a relation holds for more then one contiguous time points, we define time intervals by its start and an end point, using $between(t_1, t_2)$.
%{
%\footnotesize
%%\scriptsize
%%\begin{subequations}\label{}
%\begin{align}
%\begin{split}
%& Holds(\phi(o_1, o_2), r, between(t_1, t_2))   \leftrightarrow \\
%& \mbox{\hspace{0.3in}} \forall t \in \mathcal{T} (t_1 < t < t_2), Holds(\phi(o_1, o_2), r, at(t)) 
%\end{split}
%\end{align}
%%\end{subequations}
%}
$Occurs(\theta, at(t))$, and $Occurs(\theta, between(t_1, t_2))$ is used to denote that an event or action occurred.

\subsection{$\mathsf{\Sigma}_{\mathsf{Space}}$ -- Spatial Relations}

The theory consists of spatial relations on objects, which includes relations on \emph{topology} and \emph{extrinsic orientation} in terms of left, right, above, below relations and depth relations (distance of spatial entity from the spectator). 

\smallskip

\bul\textbf{Topology}.\quad The Region Connection Calculus (RCC) \citep{Cohn1997} is an approach to represent topological relations between regions in space. We use the RCC8 subset of the RCC, which consists of the eight base relations in $\mathcal{R}_{\mathsf{top}}$ (Figure \ref{fig:rcc8}), for representing regions of perceived objects, e.g. the projection on an object on the image plan. 
% $\phi_{\mathsf{top}}(s_i, s_j)$

{\footnotesize $\mathcal{R}_{\mathsf{top}} \equiv {\color{mathcolor}\{}\mathsf{dc}$, $\mathsf{ec}$, $\mathsf{po}$, $\mathsf{eq}$, $\mathsf{tpp}$, $\mathsf{ntpp}$, $\mathsf{tpp^{-1}}$, $\mathsf{ntpp^{-1}}{\color{mathcolor}\}}$}

\smallskip

\bul\textbf{Relative Position}.\quad 
%\bul\emph{Relative Position} 
%Relative position is represented from the viewpoint of the spectator. 
We represent the position of two spatial entities, with respect to the observer's viewpoint, using a 3-Dimensional representation that resemble Allen's interval algebra \citep{Allen1983} for each dimension, i.e. \emph{vertical}, \emph{horizontal}, and \emph{depth} (distance from the observer).  $\mathcal{R}_{pos} \equiv [\mathcal{R}_{\mathsf{pos-v}} \cup \mathcal{R}_{\mathsf{pos-h}} \cup \mathcal{R}_{\mathsf{pos-d}}]$
 %$\phi_{pos}(l_i, l_j)$

{\footnotesize $\mathcal{R}_{\mathsf{pos-v}} \equiv {\color{mathcolor}\{}\mathsf{above}$, $\mathsf{overlaps\_above}$, $\mathsf{along\_above}$, $\mathsf{vertically\_equal}$, $\mathsf{overlaps\_below}$, $\mathsf{along\_below}$, $\mathsf{below} {\color{mathcolor}\}}$}

%\smallskip

{\footnotesize $\mathcal{R}_{\mathsf{pos-h}} \equiv {\color{mathcolor}\{}\mathsf{left}$, $\mathsf{overlaps\_left}$, $\mathsf{along\_left}$, $\mathsf{horizontally\_equal}$, $\mathsf{overlaps\_right}$, $\mathsf{along\_right}$, $\mathsf{right}{\color{mathcolor}\}}$}

%\smallskip

{\footnotesize $\mathcal{R}_{\mathsf{pos-d}} \equiv {\color{mathcolor}\{}\mathsf{closer}$, $\mathsf{overlaps\_closer}$, $\mathsf{along\_closer}$, $\mathsf{distance\_equal}$, $\mathsf{overlaps\_further}$, $\mathsf{along\_further}$, $\mathsf{further}{\color{mathcolor}\}}$}

%$\mathcal{R}_{\mathsf{pos-h}} = \{left, overlaps\_left, along\_left, horizontally\_equal, overlaps\_right, along\_right, right\}$

%$\mathcal{R}_{\mathsf{pos-v}} = \{above, overlaps\_above, along\_above, vertically\_equal, overlaps\_below, along\_below, below\}$

%$\mathcal{R}_{\mathsf{pos-d}} = \{closer, overlaps\_closer, along\_closer, distance\_equal, overlaps\_further, along\_further,further\}$

%\paragraph{Intrinsic Orientation} 
%Relations for intrinsic orientation are based on the %\opram 
%relations \citep{Moratz2006}, describing the orientation (facing direction) of an robot with respect to the orientation of the observing robot.

\smallskip

\bul\textbf{Relative Distance}.\quad We represent the relative distance between two points $p_1$ and $p_2$ with respect to a third point $p_3$, using ternary relations $\mathcal{R}_{\mathsf{dist}}$. % $\phi_{\mathsf{dist}}(p_i, p_j, p_k)$.

%\smallskip

{\footnotesize $\mathcal{R}_{\mathsf{dist}} \equiv {\color{mathcolor}\{}\mathsf{closer}, \mathsf{further}, \mathsf{same}{\color{mathcolor}\}}$}

\smallskip

\bul\textbf{Relative Size}.\quad For comparison of the size of two regions we use the relations in $\mathcal{R}_{\mathsf{size}}$. % $\phi_{\mathsf{dist}}(p_i, p_j, p_k)$.

%\smallskip

{\footnotesize $\mathcal{R}_{\mathsf{dist}} \equiv {\color{mathcolor}\{}\mathsf{smaller}, \mathsf{bigger}, \mathsf{same}{\color{mathcolor}\}}$}

\subsection{$\mathsf{\Sigma}_{\mathsf{Time}}$ -- Temporal Relations}

%% ?? time point vs. moment
Temporal relations are used to represent the relationship between actions and events, e.g. one action happened before another action. We use the extensions of Allen's interval relations \citep{Allen1983} as described by \citep{Vilain1982}, i.e. these consist of relations between time \emph{points}, \emph{intervals}, and  \emph{point - interval}.
%$\phi_{\mathsf{point}}(t_i, t_j)$, $\phi_{\mathsf{interval}}(t_i, i_j)$, and $\phi_{\mathsf{point-interval}}(i_i, i_j)$.

 %point-point

%\smallskip

{\footnotesize $\mathcal{R}_{\mathsf{point}} \equiv {\color{mathcolor}\{}\timepoint$$ \mathsf{before} \timepoint$,$~~$$\timepoint \mathsf{after} \timepoint$,$~~$ $\timepoint \mathsf{equals} \timepoint{\color{mathcolor}\}}$}

%interval-interval

%\smallskip

{\footnotesize $\mathcal{R}_{\mathsf{interval}} \equiv {\color{mathcolor}\{}\mathsf{before},$ $\mathsf{after},$ $\mathsf{during},$ $\mathsf{contains},$ $\mathsf{starts},$ $\mathsf{started\_by},$ $\mathsf{finishes},$ $\mathsf{finished\_by},$ $\mathsf{overlaps},$ $\mathsf{overlapped\_by},$ $\mathsf{meets},$ $\mathsf{met\_by},$  $\mathsf{equal}{\color{mathcolor}\}}$}

%point - interval

%\smallskip

{\footnotesize $\mathcal{R}_{\mathsf{point-interval}} \equiv {\color{mathcolor}\{}\timepoint \mathsf{before}$, $\mathsf{after} \timepoint$, $\timepoint \mathsf{starts}$, $\mathsf{started\_by} \timepoint$, $\timepoint \mathsf{during}$, $\mathsf{contains}\timepoint$, $\timepoint \mathsf{finishes}$, $\mathsf{finished\_by}\timepoint$, $\timepoint \mathsf{after}$, $\mathsf{before} \timepoint {\color{mathcolor}\}}$}

%\smallskip

The relations used for temporal representation of actions and events are the union of these three, i.e. $\mathcal{R}_{\mathsf{Time}} \equiv [\mathcal{R}_{\mathsf{point}} \cup \mathcal{R}_{\mathsf{interval}} \cup \mathcal{R}_{\mathsf{point-interval}}]$.

%\begin{mydef}[Interval of Space]
%An interval  $i$ defined by its start and end points ($t_1, t_2$)  in which for all time-points between $t_1$ and $t_2$ the same relation $r$ holds is called an \emph{interval of space} 
%\end{mydef}

\subsection{$\mathsf{\Sigma}_{\mathsf{Motion}}$ -- Qualitative Spatial Dynamics}

Spatial relations holding for perceived spatial objects change as an result of motion of the individuals in the scene. To account for this, we define motion relations by making qualitative distinctions of the changes in the parameters of the objects, i.e. the distance between two depth profiles and its size. 

%Qualitative Spatial Change happening over time is represented as motion.

%\paragraph{Movement}

%changes in position

%$\mathcal{R}_{abs_move} = \{approaching, receding, static\}$

\smallskip

\bul\textbf{Relative Movement}.\quad The relative movement of pairs of spatial objects is represented in terms of changes in the distance between two points representing the objects. % $\phi_{\mathsf{move}}(p_i, p_j)$.

{\footnotesize $\mathcal{R}_{\mathsf{move}} \equiv {\color{mathcolor}\{}\mathsf{approaching}, \mathsf{receding}, \mathsf{static}{\color{mathcolor}\}}$}

\smallskip

\bul\textbf{Size Motion}.\quad 
%To represent changes in size of a single object, we use line segments.
For representing changes in size of objects, we consider relations on each dimension (\emph{horizontal}, \emph{vertical}, and \emph{depth}) separately. 
%$\phi_{\mathsf{size}}(l_i, l_j)$. 
Changes on more than one of these parameters at the same time instant can be represented by combinations of the relations.

\smallskip

{\footnotesize $\mathcal{R}_{\mathsf{size}} \equiv {\color{mathcolor}\{}\mathsf{elongating}, \mathsf{shortening}, \mathsf{static}{\color{mathcolor}\}}$}

%\bul\emph{Rate of Change}
%
%In order to represent comparison between the rate of spatial change, we use 
%
%{\footnotesize $\mathcal{R}_{rate} = \{slower$, $faster$, $same\_rate\}$}

%\subsection{Existence}
%
%The change of the existence state of a subject or object in the perceived area is denoted by the events $appearing$ and $disappearing$.
%

\section{Image Schemas of the Moving Image}\label{sec:image-schemas}
% Grounded in Visuo-Spatial Dynamics

{
\renewcommand{\arraystretch}{1.5}
\begin{table*}[htp]\scriptsize\sffamily
\caption{\emph{Image Schemas} identifiable in the literature (non-exhaustive list)}
\begin{center}
\begin{tabular}{|l||C{9cm}|}
\hline
%&\\
\textbf{SPACE} 				& \Above, \Across, \Covering, \Contact, \VerticalOrientation, \Length \\\hline
\textbf{MOTION}			& \Containment, \Path, \PathGoal, \SourcePathGoal, \Blockage, \CenterPeriphery, \Cycle, \CyclicClimax	\\\hline
\textbf{FORCE}				& \Compulsion, \Counterforce, \Diversion, \RemovalOfRestraint / \Enablement, \Attraction, \Link, \Scale	\\\hline
\textbf{BALANCE}			& \AxisBalance, \PointBalance, \TwinPanBalance, \Equilibrium	\\\hline
\textbf{TRANSFORMATION} 	& \LinearPathFromMovingObject, \PathToEndpoint, \PathToObjectMass, \MultiplexToMass, \Reflexive , \Rotation	\\\hline
\textbf{OTHERS}			& \Surface, \FullEmpty, \Merging, \Matching, \NearFar, \MassCount, \Iteration, \Object, \Splitting, \PartWhole, \Superimposition, \Process, \Collection \\\hline

\end{tabular}
\end{center}
\label{tab:image_schema}
\end{table*}
}

Image schemas have been a cornerstone in cognitive linguistics \citep{handbook-cog-linguis-geeraerts2007oxford}, and have also been investigated from the perspective of psycholinguistics, and language and cognitive development \citep{Mandler92howtobuildBaby,definingImageSchemas-2014}. Image schemas, as embodied structures  founded on experiences of interactions with the world, serve as the ideal framework for understanding and reasoning about perceived visuo-spatial dynamics, e.g., via generic conceptualisation of space, motion, force, balance, transformation, etc. Table \ref{tab:image_schema} presents a non-exhaustive list of image schemas identifiable in the literature. We formalise image schemas on individuals, objects and actions of the domain, and ground them in the spatio-temporal dynamics, as defined in Section \ref{sec:space_time_motion}, that are underling the particular schema. As examples, we focus on the spatial entities \imageschema{path}, \imageschema{CONTAINER}, \imageschema{THING}, the spatial relation \imageschema{CONTACT}, and movement relations \imageschema{MOVE, INTO, OUT OF} (these being regarded as highly important and foundational from the viewpoint of cognitive development \citep{definingImageSchemas-2014}).

\paragraph{\Containment}

%\begin{figure}[t]
%	\subfloat[][]{\includegraphics[width = \linewidth]{paper-assets/pics/drive_1_2.png}\label{sub_fig:1}}
%	
%	\subfloat[][]{\includegraphics[width = \linewidth]{paper-assets/pics/drive_1_3.png}\label{sub_fig:2}}
%	\caption{Quadrant system of the scene from The Drive, The Driver is inside the top-left quadrant, while Irene is inside the bottom-right quadrant.}
%	\label{fig:}
%\end{figure}

%\begin{figure}[t]
%	\includegraphics[width = \linewidth]{paper-assets/pics/drive_3.png}
%	
%	\caption{Quadrant system of a scene from The Drive, The Driver is inside the left quadrant, while Irene is inside the right quadrant.}
%	\label{fig:}
%\end{figure}

The \Containment schema denotes, that an object or an individual is inside of a container object.

% \scriptsize
%\begin{align}
%\begin{split}
%& Occurs(containment(entity, container), at(t)) \leftrightarrow \\
%& \mbox{\hspace{0.5in}} Holds( inside(entity, container), at(t)) \\
%\end{split}
%\end{align}
%\normalsize

%
%\begin{minipage}{\columnwidth} 
%\scriptsize
%%\footnotesize
%\begin{minted}[
%	      mathescape,
%               %linenos,
%               %numbersep=5pt,
%               gobble=0,
%               bgcolor=blue!5!white,
%               %frame=lines,
%               tabsize=1
%               %framesep=2mm
%               ]{prolog}
%containment(entity(E), container(C)) :- inside(E, C).
%\end{minted}
%\end{minipage}

\includegraphics[width=\linewidth]{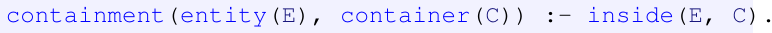}

As an example consider the following description from the film domain described in \listingref{L1}.

\sentence{Irene \spatial{occupies} the \location{right quadrant}, \spatial{while} The Driver \spatial{occupies} the  \location{left quadrant}.}

%\sentence{The Driver \spatial{occupies} the \location{top-left quadrant}, \spatial{while} Irene is \spatial{in} the \location{bottom-right quadrant}.}

In the movie example the \textsc{\sffamily entity} is a person in the film, namely The Driver, and the \textsc{\sffamily container} is a cinematographic object, the top-left quadrant, which is used to analyse the composition of the scene. We are defining the inside relation based on the involved individuals and objects, e.g. in this case we define the topological relationship between The Drivers face and the bottom-right quadrant.

%\begin{minipage}{\columnwidth} 
%\scriptsize
%%\footnotesize
%\begin{minted}[
%	      mathescape,
%               %linenos,
%               %numbersep=5pt,
%               gobble=0,
%               bgcolor=blue!5!white,
%               %frame=lines,
%               tabsize=1
%               %framesep=2mm
%               ]{prolog}
%inside(person(P), cinemat_object(quadrant(Q)) :- 
%	region(person(P), P_region),
%	region(cinemat_object(quadrant(Q)), Q_region) 
%	topology(nttp, P_region, Q_region).
%\end{minted}
%\end{minipage}

\includegraphics[width=\linewidth]{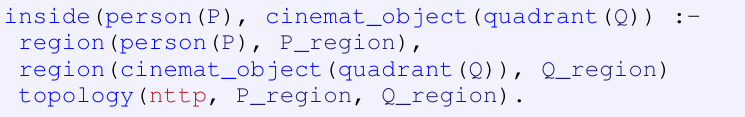}

To decide on the words to use for describing the schema, we make distinctions on the involved entities and the spatial characteristics of the scene, e.g. we use the word \emph{'occupies'}, when the person is taking up the whole space of the container, i.e. the size is bigger than a certain threshold. 
%In this example, we define an inner region of the quadrant and say, that the entity fills up the container, when the inner region of the container is a non tangential proper part of the bounding rectangle of the entity.

%\begin{minipage}{\columnwidth} 
%\scriptsize
%%\footnotesize
%\begin{minted}[
%	      mathescape,
%               %linenos,
%               %numbersep=5pt,
%               gobble=0,
%               bgcolor=blue!5!white,
%               %frame=lines,
%               tabsize=1
%               %framesep=2mm
%               ]{prolog}
%phrase(containment(E, C), [E, 'occupy', C]) :- 
%	region(person(E), E_region), 
%	region(cinemat_object(quadrant(C), C_region),
%	threshold(C_region, C_tresh),
%	size(bigger, E_region, C_tresh).
%\end{minted}
%\end{minipage}

\includegraphics[width=\linewidth]{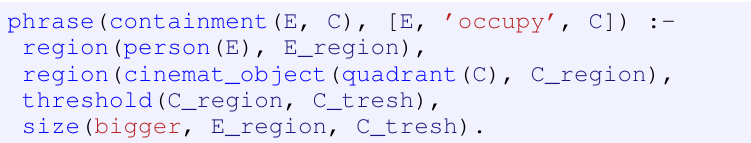}

%(topology(eq, C_region, E_region); topology(tpp, C_region, E_region);).

Similarly, we choose the word \emph{'in'}, when the person is fully contained in the quadrant.

%\paragraph{\NearFar}
%
%\medskip
%
%\begin{center}
%\colorbox{gray!18}{
%\begin{minipage}{0.96\columnwidth} \centering\footnotesize\sffamily
%
%\filmCharacter{The Driver} is \spatial{near} to the spectator seen in a \filmCinematography{close-up shot}, \spatial{while} \filmCharacter{Irene} is \spatial{far} from the spectator.
%
%\end{minipage}}
%\end{center}
%
%\medskip

%\paragraph{\PathGoal}
%
%The  \PathGoal schema involves directed movement towards a specific goal. 
%
%appraoching

\paragraph{\PathGoal and \SourcePathGoal}

% moving towards 

% subject object motion towards a goal

The \PathGoal Image Schema is used to conceptualise the movement of an \emph{object} or an \emph{individual}, towards a \emph{goal} location, on a particular \emph{path}. In this case, the path is the directed movement towards the goal.
The \SourcePathGoal Schema builds on the \PathGoal Schema by adding a source to it. Both Schemas are used to describe movement, however, in the first case, the source is not important, only the goal of the movement is of interest. Here we only describe the \SourcePathGoal Schema in more detail, as the \Path Schema is the same, without the source in it.

%The \SourcePathGoal Schema is used to conceptualise the movement of an \emph{object} or an \emph{individual}, from a \emph{source} location to a \emph{goal} location, on a particular \emph{path}.

%{
%\footnotesize
%\scriptsize
%\begin{align}
%\begin{split}
%%& (\exists (source, goal) \in \mathcal{L}) \\
%& Occurs(Source\_Path\_Goal(entity, source, path, goal), between(t_1, t_2)) \leftrightarrow \\
%%& \mbox{\hspace{0.5in}} (\exists t_1, t_2) (t_1 < t_2) \wedge\\
%& \mbox{\hspace{0.5in}}  Holds(at\_location(entity, source), true, at(t_1)) \wedge\\
%& \mbox{\hspace{0.5in}}  Holds(at\_location(entity, goal), true, at(t_2)) \wedge\\
%& \mbox{\hspace{0.5in}}  Occurs(move(entity, path), between(t_1, t_2))\\
%\end{split}
%\end{align}
%}

%\begin{minipage}{\columnwidth} 
%\scriptsize
%%\footnotesize
%\begin{minted}[
%	      mathescape,
%               %linenos,
%               %numbersep=5pt,
%               gobble=0,
%               bgcolor=blue!5!white,
%               %frame=lines,
%               tabsize=1
%               %framesep=2mm
%               ]{prolog}
%source_path_goal(Trajector, Source, Path, Goal) :-
%	entity(Trajector), location(Source), location(Goal),
%	path(Path, Source, Goal),
%	at_location(Trajector, Source, at_time(T_1)), 
%	at_location(Trajector, Goal, at_time(T_2)),
%	move(Trajector, Path, between(T_1, T_2)).
%\end{minted}
%\end{minipage}

\includegraphics[width=\linewidth]{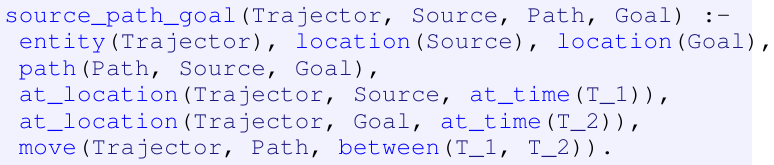}

In the way finding analysis one example of the \SourcePathGoal schema is when a description of the path a subject was walking is generated.

\sentence{ Barbara  \action{walks} \spatial{from} the \location{emergency}, \spatial{through} the \location{atrium lobby} \spatial{to} the \location{blue elevators}.}

Another example is when a descriptions of a subjects eye movement is generated from the eye tracking experiment. 

\begin{center}
\colorbox{gray!7}{
\begin{minipage}{0.96\columnwidth} \centering\sffamily\footnotesize
Barbaras eyes \action{move} \spatial{from} the \location{emergency sign},\\\spatial{over} the \location{exit sign} \spatial{to} the \location{elevator sign}.
\end{minipage}}
\end{center}

\smallskip

In both of these sentences there is a moving entity, the \emph{trajector}, a \emph{source} and a \emph{goal} location, and a \emph{path} connecting the source and the goal. In the first sentence it is Barbara who is moving, while in the second sentence Barbaras eyes are moving. Based on the different spatial entities involved in the movement, we need different definitions of locations, path, and the moving actions. In the way finding domain, a subject is at a location when the position of the person upon a 2-dimensional floorplan is inside the region denoting the location, e.g. a room, a corridor, or any spatial artefact describing a region in the floorplan.

%\begin{minipage}{\columnwidth} 
%\scriptsize
%%\footnotesize
%\begin{minted}[
%	      mathescape,
%               %linenos,
%               %numbersep=5pt,
%               gobble=0,
%               bgcolor=blue!5!white,
%               %frame=lines,
%               tabsize=1
%               %framesep=2mm
%               ]{prolog}
%
%at_location(Subject, Location) :- 
%	person(Subject), room(Location),
%	position(Subject, S_pos), region(Location, L_reg),
%	topology(ntpp, S_pos, Loc_reg).
%	
%		
%\end{minted}
%\end{minipage}

\includegraphics[width=\linewidth]{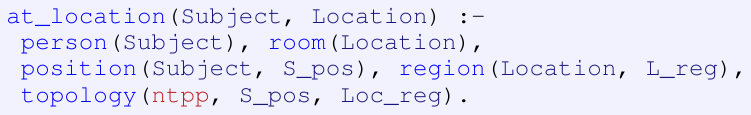}

Possible paths between the locations of a floorplan are  represented by a topological route graph, on which the subject is walking.

%\begin{minipage}{\columnwidth} 
%\scriptsize
%%\footnotesize
%\begin{minted}[
%	      mathescape,
%               %linenos,
%               %numbersep=5pt,
%               gobble=0,
%               bgcolor=blue!5!white,
%               %frame=lines,
%               tabsize=1
%               %framesep=2mm
%               ]{prolog}
%move(person(Subject), Path) :-
%	action(movement(walk), Subject, Path),
%	movement(approaching, Subject, Goal).
%\end{minted}
%\end{minipage}

\includegraphics[width=\linewidth]{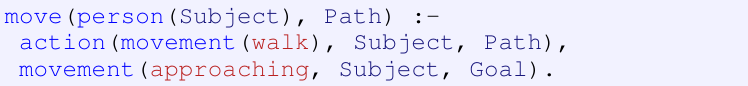}

%When generating summaries of the eye tracking experiments, the locations are objects, e.g. signs, the subject looks at, while walking through the environment and the movement refers to the subjects eye movement.
For generating language, we have to take the type of the trajector into account, as well as the involved movement and the locations, e.g. the eyes are moving \emph{'over'} some objects, but Barbara moves \emph{'trough'} the corridor.

%\begin{minipage}{\columnwidth} 
%\scriptsize
%%\footnotesize
%\begin{minted}[
%	      mathescape,
%               %linenos,
%               %numbersep=5pt,
%               gobble=0,
%               bgcolor=blue!5!white,
%               %frame=lines,
%               tabsize=1
%               %framesep=2mm
%               ]{prolog}
%phrase([Action, 'from', Source, 'through', Path, 'to', Goal]) :-
%action(movement(Action)), 
%room(Source)
%\end{minted}
%\end{minipage}

\paragraph{\Attraction}

The \Attraction schema is expressing a force by which an entity is attracted.

%{
%\footnotesize
%\scriptsize
%\begin{subequations}\label{}
%\begin{align}
%\begin{split}
%& Occurs(attraction(subject, entity), between(t_1, t_2))  \leftarrow \\
%& \mbox{\hspace{0.5in}}  Occurs(force(attraction(subject, entity), between(t_1, t_2))\\
%\end{split}
%\end{align}
%\end{subequations}\normalsize
%}
%
%\begin{minipage}{\columnwidth} 
%\scriptsize
%%\footnotesize
%\begin{minted}[
%	      mathescape,
%               %linenos,
%               %numbersep=5pt,
%               gobble=0,
%               bgcolor=blue!5!white,
%               %frame=lines,
%               tabsize=1
%               %framesep=2mm
%               ]{prolog}
%attraction(Subject, Entity) :-
%	entity(Subject), entity(Entity),
%	force(attraction, Subject, Entity).	
%\end{minted}
%\end{minipage}

\includegraphics[width=\linewidth]{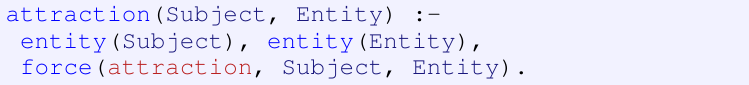}

An example for \Attraction is the eye tracking experiment, when the attention of a subject is attracted by some object in the environment.

\sentence{While walking \spatial{through} the \location{hallway}, Barbaras attention is {\color{blue}attracted} by the \location{outside view}.}

In this case the entity is Barbara's attention which is represented by the eye tracking data, and it is attracted by the force, the outside view applies on it.
We define attraction by the fact, that the gaze position of \textsf{Barbara} has been on the outside for a substantial amount of time, however, this definition can be adapted to the needs of domain experts, e.g. architects who want to know what are the things that grab the attention of people in a building.

%phrase([Obj1, 'is', 'attracted by', Obj2],  between(T1, T2))

%\section{Declarative Natural Language Generation}
\section{From Perceptual Narratives to Natural Language}\label{sec:language-generation}
The design and implementation of the natural language generation component has been driven by three key developmental goals: (1) ensuring support for, and uniformity with respect to the (deep) representational semantics of space and motion relations etc (Section \ref{sec:space_time_motion}); (2) development of modular, yet tightly integrated set of components that can be easily used within the state-of-the-art (constraint) logic programming family of KR methods; and (3) providing seamless integration capabilities within hybrid AI and computational cognition systems.

\subsection*{System Overview (NL Generation)}
The overall pipeline of the language generation component follows a standard natural language generation system architecture \citep{ReiterDale2000,bateman2003NLG}. Figure \ref{fig:architecture-language-generation} illustrates the system architecture encompassing the typical  stages of content determination \& result structuring, linguistic \& syntactic realisation, and syntax tree \& sentence generation.

\begin{figure*}[t]
    \centering
    \includegraphics[width=0.94\textwidth]{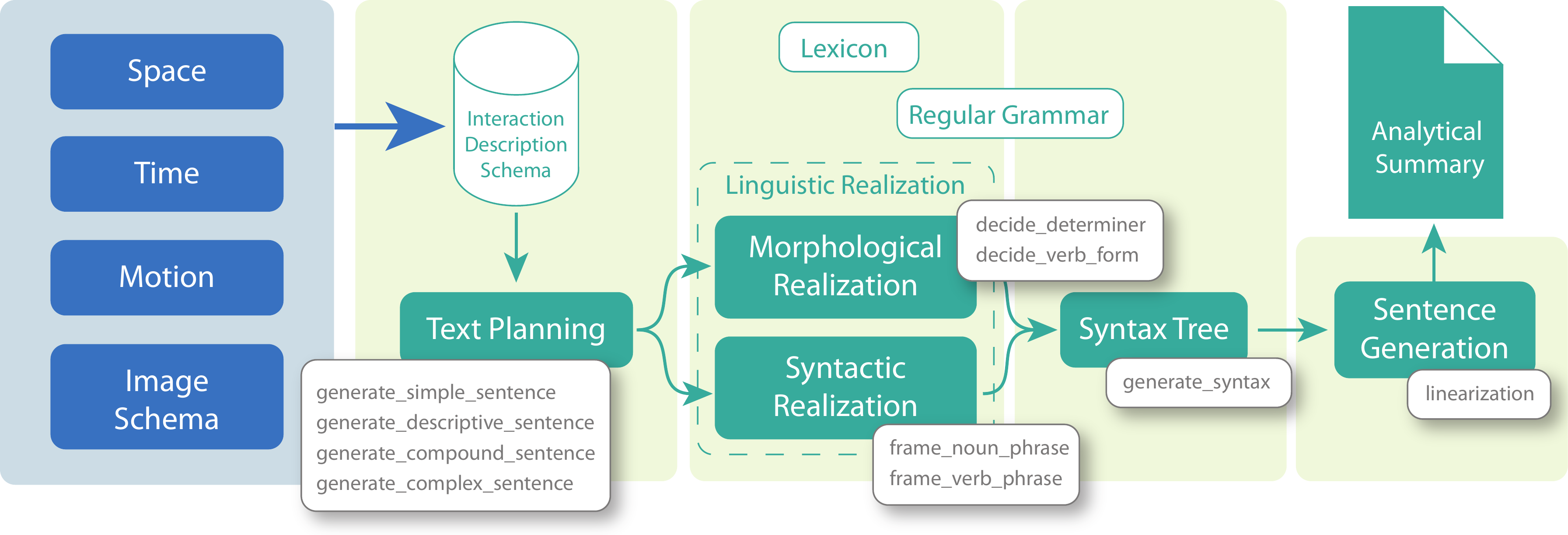}
    \caption{From Perceptual Narratives to Natural Language}
    \label{fig:architecture-language-generation}
\end{figure*}

\subsubsection*{S1.\quad Input -- Interaction Description Schema}
Interfacing with the language generator is possible with a generic (activity-theoretic) Interaction Description Schema (IDS) that is founded on the ontology of the (declarative) perceptual narrative, and a general set of constructs to introduce the domain-specific vocabulary. Instances of the IDS constitute the domain-specific input data for the generator.

\subsubsection*{S2.\quad Syntax Tree and Sentence Generation}
The generator consists of sub-modules concerned with input IDS instance to text planning, morphological \& syntanctic realisation, and syntax tree \& sentence generation. Currently, the generator functions in a \emph{single interaction} mode where each invocation of the system (with an input instance of the IDS) produces a single sentence in order to produce spatio-temporal domain-based text. The morphological and syntactic realisation module brings in assertions of detailed grammatical knowledge and the lexicon that needs to be encapsulated for morpohological realisation; this encompasses aspects such as noun and verb categories, spatial relations and locations; part of speech identification is also performed at this stage, including determiner and adjective selection, selection of verb and tense etc. The parts of speech identified by the morph analyser taken together with context free grammar rules for simple, complex, and compound sentence constructions are used for syntactic realisation, and sentence generation.

%\begin{minipage}{\columnwidth} 
%\scriptsize
%%\footnotesize
%\begin{minted}[
%	      mathescape,
%               %linenos,
%               %numbersep=5pt,
%               gobble=0,
%               bgcolor=blue!5!white,
%               %frame=lines,
%               fontfamily=fi4,
%               tabsize=1
%               %framesep=2mm
%               ]{prolog}
%generateSentence([ [], 'Barbara', [], focus, on, 
%	'Drinking Water', sign, at, corridor, end ], SENTENCE).
%\end{minted}
%\end{minipage}

%\begin{minipage}{\columnwidth} 
%\scriptsize
%%\footnotesize
%\begin{minted}[
%	      mathescape,
%%               linenos,
%               numbersep=5pt,
%               gobble=0,
%               bgcolor=black!5!white,
%               %frame=lines,
%               %fontfamily=fi4,
%               tabsize=1
%               %framesep=2mm
%               ]{prolog}
% 1.	Properties (Interaction Subject) - ADJECTIVE
% 2.	Name of the subject in the interaction - NOUN
% 3.	Analysis of the motion identified - ADVERB
% 4.	Name of motion identified - MAIN VERB
% 5. 	ObjectList [Relation between the two 
%				consecutive results - PREPOSITION,
%				Properties of the object- ADJECTIVE,
%				Name of the object] - NOUN
% 6.	Expected Tense of Action - TENSE
%\end{minted}
%\end{minipage}

%\subsubsection*{\small \mylabel{M3}\quad \textsc{Syntactic realisation}}

%\subsubsection*{\small \mylabel{S3}\quad \textsc{Syntax tree and sentence generation}}
%\subsubsection*{\small \mylabel{M5}\quad \textsc{Sentence generation}}

\subsection*{Language Generation (Done Declaratively)}
Each aspect of generation process, be it at a factual level (grammar, lexicon, input data) or at a process level (realisation, syntax tree generation) is fully \emph{declarative} (to the extent possible in logic programming) and \emph{elaboration tolerant} (i.e., addition or removal or facts \& rules,  constraints etc does not break down the generation process). An important consequence of this level of declarativeness is that a query can work both ways: from input data to syntax tree to sentence, or from a sentence back to its syntax tree and linguistic decomposition wrt. to a specific lexicon.

%\begin{enumerate}
%	\item Text planning
%
%	\item Linguistic realisation
%	
%	\item Syntactic realisation
%	
%	\item Syntax tree generation
%	
%	\item Sentence generation
%\end{enumerate}

%\begin{table}[htdp]
%\caption{Time taken for generating summaries (in ms)}
%\begin{center}\footnotesize
%\begin{tabular}{l|rrr}
%\textbf{Tense} & \textbf{Avg.} & \textbf{Min.} & \textbf{Max.} \\\hline
%Simple & 77.8 & 70 & 96\\
%Continous & 84.48 & 73 & 99\\
%\end{tabular}
%\end{center}
%\label{default}
%\end{table}%
%
%
%\begin{table}[htdp]
%\caption{}
%\begin{center}\footnotesize
%\begin{tabular}{l|r}
%\textbf{Type} & \textbf{Time}  \\\hline
%simple  & 0,52 \\
%compound  & 1,23 \\
%complex  & 1,32 \\
%\end{tabular}
%\end{center}
%\label{default}
%\end{table}%

%\begin{table}[htdp]
%\caption{Time taken for generating summaries (in ms)}
%\begin{center}\scriptsize
%\begin{subtable}[c]{0.64\linewidth}
%\begin{tabular}{l|rrr}
%\textbf{Tense} & \textbf{Avg.} & \textbf{Min.} & \textbf{Max.} \\\hline
%Simple & 77.8 & 70 & 96\\
%Continous & 84.48 & 73 & 99\\
%\end{tabular}
%\subcaption{}
%\end{subtable}
%\begin{subtable}[c]{0.34\linewidth}
%\begin{tabular}{l|r}
%\textbf{Type} & \textbf{Time}  \\\hline
%simple  & 0,52 \\
%compound  & 1,23 \\
%complex  & 1,32 \\
%\end{tabular}
%\subcaption{}
%\end{subtable}
%\end{center}
%\label{default}
%\end{table}%

\subsection*{Empirical Evaluation of Language Generation}
We tested the language generation component with data for $25$ subjects, $500$ IDS instances, and $53$ domain facts ({\small using an Intel Core i7-3630QM CPU @ 2.40GHz x 8}). We generated summaries in simple/continuous  present, past, future respectively for all IDS instances. Table (\ref{tab:evaluation}): {(a)}. average of 20 interactions, on an average 26.2 sentences / summary, with 17.6 tokens as the average length / sentence; (b)  generated 100 sentences for simple, compound, and complex types reflecting the average sentence generation time.

\begin{table}[htp]
\caption{Time  (in ms) for (a) summaries, (b) sentences}
\begin{center}\footnotesize
\begin{tabular}{ccc}
\begin{tabular}{l|rrr}
\textbf{Tense} & \textbf{Avg.} & \textbf{Min.} & \textbf{Max.} \\\hline
simple & 77.8 & 70 & 96\\
continous & 84.48 & 73 & 99\\
\end{tabular}
&
\hspace{0.5in}
&
\begin{tabular}{l|r}
\textbf{Type} & \textbf{Time}  \\\hline
simple  & 0,52 \\
compound  & 1,23 \\
complex  & 1,32 \\
\end{tabular}\\
(a)&&(b)
\end{tabular}
\label{tab:evaluation}
\end{center}
\label{default}
\end{table}%

\section{\textsc{Discussion and Related Work}}\label{sec:discussion-and-relwork}
Cognitive vision as an area of research has already gained prominence, with several recent initiatives addressing the topic from the perspectives of language, logic, and artificial intelligence \citep{Vernon2006,Vernon2008,ILP-2011-Dubba,CMN-2013-Bhatt,suchan-embodied-pricai2014,JAIR-ILP-Leeds-Bremen-2015}. There has also been an increased interest from the computer vision community to synergise with cognitively motivated methods for language grounding and inference with visual imagery  \citep{Li-Fei-Fei-2015-deep-cvpr,compositional-grounding-jair2015}. This paper has not attempted to present advances in basic computer vision research; in general, this is not the agenda of our research even outside the scope of this paper. The low-level visual  processing algorithms that we utilise are founded in state-of-the-art outcomes from the computer vision community for detection and tracking of \emph{people, objects, and motion} \citep{Canny1986,Lucas1981,Viola2001,Dalal2005}.\footnote{For instance, we analyse motion in a scene sparse and dense optical flow \citep{Lucas1981,Farneback2003}, detecting faces using cascades of features \citep{Viola2001}, detecting humans using histograms of oriented gradients \citep{Dalal2005}.} On the language front, the number of research projects addressing natural language generation systems \citep{ReiterDale2000,bateman2003NLG} is overwhelming; there exist a plethora of projects and initiatives focussing on language generation in general or specific contexts, candidate examples being the works in the context of \emph{weather report} generation \citep{NLG-weather-1994,NLG-weather-2014}, \emph{Pollen} forecasts \citep{NLG-ST-Pollen-2006}.\footnote{We have been unable to locate a fitting \& comparable spatio-temporal feature sensitive language generation module for open-source usage. We will disseminate our language generation component as an open-source \textsc{Prolog} library.} Our focus on the (declarative) language generation component of the framework of this paper (Section \ref{sec:language-generation}) has been on the use of ``deep semantics'' for space and motion, and to have a unified framework --with each aspect of the embodied perception grounding framework-- fully implemented within constraint logic programming.

Our research is motivated by computational cognitive systems concerned with interpreting multimodal dynamic perceptual input; in this context, we believe that it is essential to build systematic methods and tools for {embodied visuo-spatial conception, formalisation, and computation} with primitives of space and motion. Toward this, this paper has developed a declarative framework for embodied grounding and natural language based analytical summarisation of the moving image; the implemented model consists of modularly built components for logic-based representation and reasoning about qualitative and linguistically motivated abstractions about space, motion, and image schemas.  Our model and approach can directly provide the foundations that are needed for the development of novel assistive technologies in areas where high-level qualitative analysis and sensemaking  \citep{Bhatt-Schultz-Freksa:2013,Bhatt-sensemaking-narrative-2013} of dynamic visuo-spatial imagery is central.

\section*{Acknowledgements}
We acknowledge the contributions of DesignSpace members Saurabh Goyal, Giulio Carducci, John Sutton, and Vasiliki Kondyli in supporting developmental, design, experimentation, and expert (qualitative) analysis tasks.

\bibliographystyle{abbrvnat}
%\bibliography{bibs/film-bib,bibs/cognitive-vision,bibs/NLG-bib,bibs/image-schema,bibs/space-time-motion,bibs/computer-vision}

\end{document}